%% file: main.tex
\documentclass{article} 
\usepackage[OT1]{fontenc} 
\usepackage{iclr2027_conference,times}

\input{math_commands.tex}

\usepackage{amsmath}
\usepackage{amssymb}
\usepackage{booktabs}
\usepackage{graphicx}
\usepackage{placeins}
\usepackage{microtype}
\usepackage{xcolor}
\usepackage{colortbl}
\usepackage{hyperref}
\usepackage{url}
\usepackage{wrapfig}
\usepackage{tabularx}
\usepackage{textcomp}

\title{Visual Parallel Search: Learning to Search High-Resolution Images with Parallel Tile Inspection and Adaptive Zoom}

\newcommand{\hkuaff}{\ensuremath{\spadesuit}}
\newcommand{\huaweiaff}{\ensuremath{\heartsuit}}
\newcommand{\cityuaff}{\ensuremath{\clubsuit}}
\newcommand{\hustaff}{\ensuremath{\diamondsuit}}

\author{
\textbf{%
Xijia Tao\textsuperscript{\hkuaff,*} \quad
Yihua Teng\textsuperscript{\huaweiaff,*} \quad
Xinyu Fu\textsuperscript{\huaweiaff} \quad
Cheng Gong\textsuperscript{\cityuaff}} \\
\textbf{%
Ziru Liu\textsuperscript{\huaweiaff} \quad
Xudong Xie\textsuperscript{\hustaff} \quad
Rui Liu\textsuperscript{\huaweiaff,\textdagger} \quad
Lingpeng Kong\textsuperscript{\hkuaff,\textdagger}} \\
\normalfont
\textsuperscript{\hkuaff}The University of Hong Kong \quad
\textsuperscript{\huaweiaff}Huawei Research \\
\textsuperscript{\cityuaff}City University of Hong Kong \quad
\textsuperscript{\hustaff}Huazhong University of Science and Technology \\
\textsuperscript{*}Equal contribution \quad
\textsuperscript{\textdagger}Corresponding authors
}

\definecolor{methodpurple}{HTML}{8B32E8}
\definecolor{methodgreen}{HTML}{00A35C}

\newcommand{\method}{\textcolor{methodpurple}{\textsc{VPS}}}

\newcommand{\tbd}{--}
\newcommand{\best}[1]{\textbf{\boldmath #1}}

\AddToHook{env/table/begin}{\setlength{\belowcaptionskip}{7pt}}
\AddToHook{env/table*/begin}{\setlength{\belowcaptionskip}{7pt}}
\definecolor{tablefocus}{HTML}{F2EDF9}
\definecolor{tablepositive}{HTML}{005F86}
\definecolor{tablenegative}{HTML}{A34818}
\newcommand{\posdelta}[1]{\textcolor{tablepositive}{\ensuremath{#1}}}
\newcommand{\negdelta}[1]{\textcolor{tablenegative}{\ensuremath{#1}}}

\iclrfinalcopy

\begin{document}

\maketitle
\fancyhead{}

\input{sections/0_abstract}
\input{sections/1_introduction}
\input{sections/3_method}
\FloatBarrier
\input{sections/4_data}
\input{sections/5_experiments}
\input{sections/6_results}
\FloatBarrier
\input{sections/8_conclusion}

\newpage







\bibliography{references}
\bibliographystyle{iclr2027_conference}

\clearpage
\appendix
\input{sections/2_related_work}
\input{sections/7_analysis}
\input{sections/9_appendix}

\end{document}

%% file: math_commands.tex
\usepackage{amsmath,amsfonts,bm}

\def\eqref#1{equation~\ref{#1}}

\def\1{\bm{1}}

\DeclareMathAlphabet{\mathsfit}{\encodingdefault}{\sfdefault}{m}{sl}
\SetMathAlphabet{\mathsfit}{bold}{\encodingdefault}{\sfdefault}{bx}{n}



%% file: sections/0_abstract.tex
\begin{abstract}
High-resolution visual question answering often fails because a multimodal model does not acquire the small, spatially localized evidence needed to answer a question.  Sequential zooming can recover detail, but it asks the main model to choose a region before obtaining a reliable overview.  We introduce \method{}, a visual parallel-search framework in which a main agent first invokes \texttt{grid\_search} to inspect image tiles in parallel with question-conditioned sub-agents, and then adaptively invokes \texttt{zoom\_in} on a precise or merged region.  The same interface supports both training-free inference and post-training of the main and sub-agents.  Across five benchmark splits and three model sizes, \method{} improves mean accuracy over dedicated zoom-only search in 14 of 15 same-model comparisons, with gains up to $8.0$ points and especially strong improvements for smaller main models.  ZoomBench retains an approximately $3.2$-point gain at every tested size.  We further develop a supervision pipeline with hint-free verification and a paired role-specific GRPO surrogate for learning the controller and tile-reader roles.  SFT improves observed accuracy on all five benchmark splits, including a $4.17$-point gain on HR-Bench 4K.  Role-specific RL further reshapes search behavior: main-only RL reduces mean tool use from $2.65$ to $2.11$ with similar pass@1 in an internal four-response evaluation, while external accuracy changes are mixed.  Joint training reveals an asymmetry between local evidence reading and global search control.  Together, these results support \method{} as an effective inference-time scaffold and a trainable decomposition for visual evidence acquisition. 
Project page: \url{https://xijia-tao.github.io/vps/}
\end{abstract}

%% file: sections/1_introduction.tex
\section{Introduction}
\label{sec:introduction}

Multimodal language models can reason about an image only after the relevant visual evidence reaches their context at a usable resolution.  This creates a basic acquisition problem on high-resolution images: a label may occupy a few pixels, a chart answer may require a local reading, and a relational question may require evidence from multiple distant regions.  Feeding a globally resized image can erase the answer, while repeatedly zooming without a visual prior turns perception into an expensive sequential search.

Existing crop-and-zoom systems demonstrate that active visual inspection is valuable, but they commonly organize search around fixed geometric patches or ask one agent to propose a region from the initial thumbnail \citep{wu2023vstar,wang2025dc2,shen2025zoomeye}.  Fixed partitions can split an object or table across boundaries, and a single sequential agent must trade coverage for local detail.  More importantly, these designs offer a weak interface for learning: local perception, region selection, evidence aggregation, and final answering are entangled in one long trajectory.

We study \emph{visual parallel search}: a main agent delegates question-conditioned inspection of a regular grid to multiple tile readers, receives compact observations from all tiles, and uses them to decide whether and where to zoom.  Our framework, \method{}, consists of two core operations.  \texttt{grid\_search} provides a high-recall overview by inspecting tiles in parallel; \texttt{zoom\_in} then recovers fine detail from a selected region or from the union of neighboring tiles.  The main agent remains responsible for search control and answer synthesis, while the sub-agent specializes in local evidence perception.

V* uses guided visual search, DC$^2$ recursively partitions images into local textual evidence, and ZoomEye explores a hierarchy of regions \citep{wu2023vstar,wang2025dc2,shen2025zoomeye}.  \method{} couples a regular spatial decomposition with parallel readers and controller-directed region merging.

Learned orchestration connects this design to tool routing in ToolOrchestra \citep{su2025toolorchestra}, task-specific executors in AOrchestra \citep{ruan2026aorchestra}, and parallel dispatch in ParaManager and Orchestra-o1 \citep{yuan2026paramanager,zhang2026orchestrao1}.  \method{} uses homogeneous tile readers and a fixed spatial decomposition, connecting each worker report to an image region.  This makes local perception separately supervisable while preserving controller decisions about coverage, evidence aggregation, and follow-up inspection.  Appendix~\ref{sec:related_work} provides the broader comparison.

This decomposition is useful beyond inference.  Local tile judgments can be supervised with region annotations and verified independently, while the main policy can be trained on end-task reward.  We therefore develop both a supervised data recipe and a paired main/sub-agent reinforcement-learning recipe.  The latter treats the two roles as related but distinct GRPO tasks \citep{shao2024deepseekmath}: responses are compared only within groups sharing an identical prompt, and task losses are combined with an explicit weight.  We also distinguish this practical, role-masked surrogate from the full joint-system policy gradient, which would additionally propagate final-answer reward through the sampled sub-agent reports.

\begin{figure}[t]
\centering
\includegraphics[width=\linewidth]{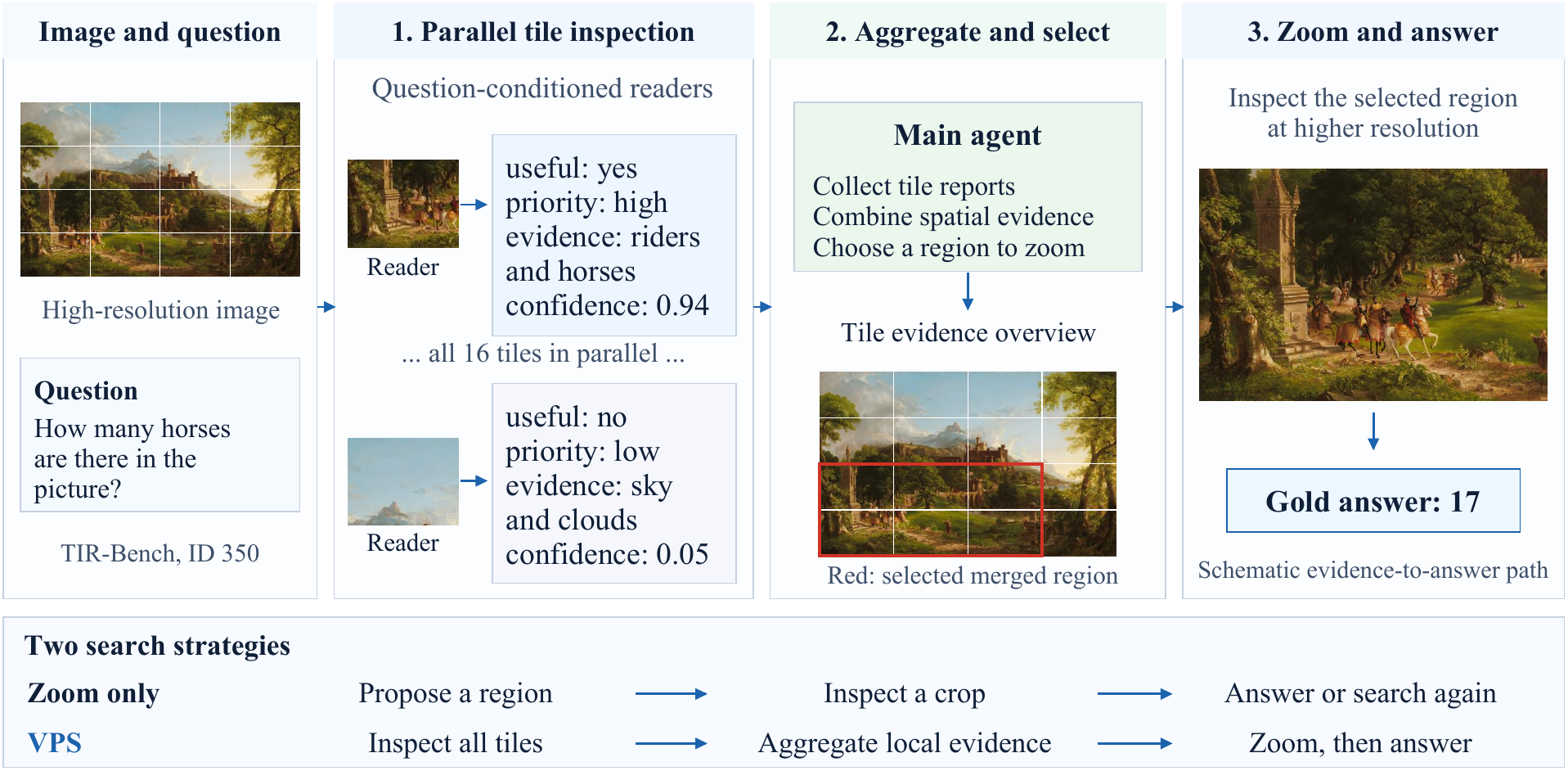}
\caption{\method{} combines parallel tile inspection with adaptive zoom. The main agent aggregates local evidence and selects a precise or merged region for inspection. The image and question are from TIR-Bench visual search, sample ID \texttt{350} (gold answer: 17; \citealp{li2025tirbench}). Reports, selected regions, and the comparison strip illustrate the mechanism rather than reproduce a measured rollout.}
\label{fig:teaser}
\end{figure}

Our experiments compare \method{} with dedicated zoom-only search across model sizes and image resolutions, then examine how SFT and RL change tool behavior.  Mean accuracy improves in 14 of 15 same-model cells, and the advantage is generally larger for 4B and 9B main agents than for a 27B main agent; at 27B, only ZoomBench and HR-Bench 4K exceed replicate spread.  Supervised fine-tuning raises observed accuracy on all five evaluated benchmark splits, with a matched unseeded HR-Bench 4K gain of $4.17$ points.  Role-specific GRPO further adapts the search policy: main-only RL reduces mean tool calls from $2.65$ to $2.11$ with similar pass@1 in an internal four-response evaluation, while joint training exposes an asymmetry between local evidence reading and global search control.  Deployment-role comparisons show that the joint-trained main agent takes longer trajectories and completes fewer searches.

Our contributions are:\nopagebreak[4]
\begin{itemize}
    \item a minimal parallel-search harness that combines question-conditioned tile inspection with adaptive single- or multi-tile zoom;
    \item a systematic comparison against dedicated zoom-only search across three model sizes, five benchmark splits, and replicated evaluations;
    \item a data recipe for main-agent trajectories and hint-filtered sub-agent ROI supervision; and
    \item a paired role-specific GRPO surrogate with explicit gradient accounting, and an empirical analysis of tool use and the different effects of training the controller and tile reader.
\end{itemize}

%% file: sections/3_method.tex
\section{Visual Parallel Search}
\label{sec:method}

\subsection{Problem Setup}
Given an image $I$, question $q$, and answer $a^*$, a visual-search policy produces a trajectory $\tau=(o_0,u_1,o_1,\ldots,u_T,\hat a)$, where $u_t$ is a tool call or final-answer action and $o_t$ is the resulting observation.  The objective is to maximize answer quality subject to a bounded number of turns and visual observations.  We factor the policy into a main agent $\pi_\theta^{\mathrm{main}}$, which controls the trajectory, and a tile reader $\pi_\theta^{\mathrm{sub}}$, which assesses local evidence.  The two roles may share model weights but use different prompts and response masks.

\subsection{Parallel Tile Inspection}
\label{sec:grid_search}
The main agent calls \texttt{grid\_search}$(I,n)$ with $n\in\{3,\ldots,8\}$.  The tool partitions the current image into an $n\times n$ grid and dispatches each tile in parallel to the sub-agent.  For tile $j$, the sub-agent returns a compact record
\begin{equation}
o_j=(\texttt{useful},\texttt{priority},\texttt{evidence},
      \texttt{candidate\_answer},\texttt{bbox},\texttt{confidence}).
\end{equation}
The tile reader sees the tile, question, tile identifier, and tile coordinates, but not the gold ROI or the other tiles.  The harness aggregates the records into a model-visible overview.  It does not choose a winning tile: selection remains an action of the main agent.

\subsection{Adaptive Zoom and Region Aggregation}
The main agent calls \texttt{zoom\_in}$(I,b)$ with either one normalized box $b=[x_1,y_1,x_2,y_2]$ or a list of neighboring boxes.  In the latter case the tool returns the smallest enclosing crop, allowing the model to repair grid-boundary fragmentation and inspect relational evidence.  This division of labor gives grid search high recall and zoom high spatial precision.  The current system performs one layer of grid search; recursive search is left to future work.

\subsection{Agent Loop}
Starting from a globally resized view, the main agent may (i) answer, (ii) scan a grid, or (iii) zoom into the original-resolution image.  Observations are appended to the trajectory until the model answers or reaches the turn budget.  A final-turn instruction requests an answer when the budget is exhausted.  We record tool calls, crop lineage, original-image coordinates, token usage, and termination state for evaluation and training.

\subsection{Supervised Post-Training}
\label{sec:sft}
We train the main role on successful tool-use trajectories and the sub role on verified tile judgments.  Main-agent examples teach when to search, how to interpret the grid report, when to merge regions, and when to stop.  Sub-agent examples teach schema validity, local usefulness, evidence description, priority, candidate answering, and evidence localization.  Section~\ref{sec:data} describes how we prevent ROI hints from leaking into the deployed tile-reader prompt.
The selected SFT policy is the cold-start checkpoint for orchestration RL; RL improvements are measured relative to this SFT policy rather than directly from the pretrained backbone.

\begin{figure}[tbp]
\centering
\includegraphics[width=\linewidth]{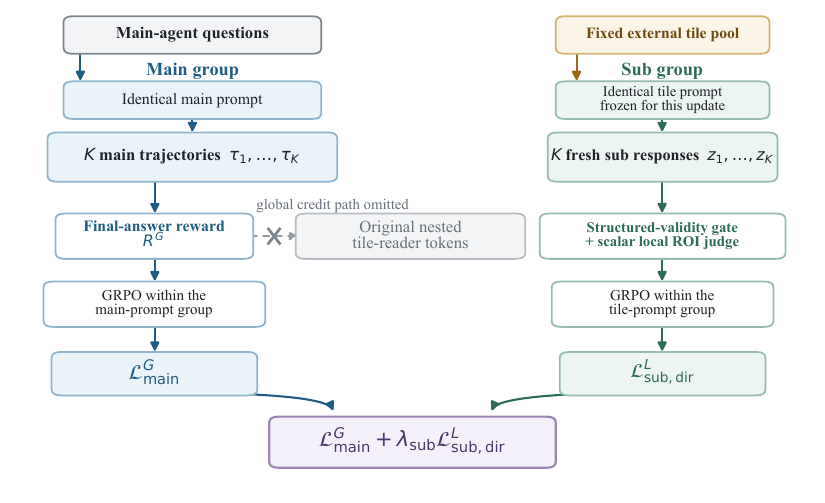}
\caption{Paired role-specific GRPO. Main questions and fixed external tile prompts form separate groups of $K$ responses, each normalized within an identical prompt. Main tokens receive final-answer supervision; fresh tile responses receive local supervision. The two losses update shared weights. The crossed path marks omitted global credit to nested tile-reader tokens.}

\label{fig:paired_grpo}
\end{figure}

\subsection{Paired Role-Specific GRPO}
\label{sec:rl}
The system is a joint stochastic policy when both roles are trainable.  Let $\mathcal I_{\mathrm{main}}(\tau)$ and $\mathcal I_{\mathrm{sub}}(\tau)$ index tokens sampled by the main call and by the nested tile-reader calls.  If tool transitions and aggregation contain no trainable parameters, the score-function gradient of final-answer return $R_G$ contains both paths:
\begin{equation}
\nabla J_G=\mathbb E\!\left[(R_G-b)\left(
\sum_{u\in\mathcal I_{\mathrm{main}}}\nabla\log\pi^{\mathrm{main}}(a_u\mid h_u)
+\sum_{v\in\mathcal I_{\mathrm{sub}}}\nabla\log\pi^{\mathrm{sub}}(a_v\mid h_v)
\right)\right].
\label{eq:joint_score}
\end{equation}
The second term exists even when each tile reader is sampled only once.  Broadcasting one terminal return to many parallel reports, however, gives diffuse, high-variance credit: a correct report can be penalized when another report or the main agent fails, and an incorrect report can be rewarded when another branch recovers.

Our paired surrogate combines two separately sampled tasks.  For each main-agent prompt $i$, we sample $K$ trajectories under that identical prompt, giving group $G_i^{\mathrm{main}}$.  In the reported joint runs, sub-agent prompts come from a fixed external tile dataset.  For each tile prompt $j$, we sample $K$ responses to that same prompt, giving $G_j^{\mathrm{sub}}$.  Main trajectories do not supply the directly supervised tile examples in these runs.  Figure~\ref{fig:paired_grpo} illustrates these two training branches.  Main and sub groups receive distinct identifiers and are normalized separately:
\begin{align}
A_{i,k}^{\mathrm{main}} &= \operatorname{GRPO}(R_{i,1:K}^{\mathrm{main}})_k,\\
A_{j,k}^{\mathrm{sub}}  &= \operatorname{GRPO}(R_{j,1:K}^{\mathrm{sub}})_k.
\end{align}
We optimize
\begin{equation}
\mathcal{L}_{\mathrm{paired}}
=\mathcal{L}_{\mathrm{main}}^{G}
+\lambda_{\mathrm{sub}}\mathcal{L}_{\mathrm{sub,dir}}^{L},
\label{eq:joint_objective}
\end{equation}
where $\lambda_{\mathrm{sub}}$ denotes the relative local-loss weight.  The main loss is applied only to main-generated tokens; copied tool observations and the original nested sub-agent outputs receive no global-return policy loss.  The local loss is applied only to newly sampled sub-agent responses in the fixed tile-prompt batch.  Thus Eq.~\ref{eq:joint_objective} omits the second score term in Eq.~\ref{eq:joint_score}.  The external tile distribution has no main-policy occupancy derivative.  If tiles instead come from on-policy grid calls, freezing their prompts for the update omits that additional derivative.  With shared weights the combined objective is a biased, lower-variance semi-gradient relative to the full joint return.  With an independently frozen deployed sub-agent, the main term instead targets the controller objective conditional on that fixed tool, up to the usual GRPO/PPO surrogate approximations \citep{shao2024deepseekmath,schulman2017ppo}; after deploying a separately trained replacement sub-agent, main trajectories must be recollected before the next controller update.

Main groups receive final-answer supervision; optional cost shaping is a separate intervention.  For sub groups, contemporaneous records describe a structured-output gate followed by a scalar multimodal-judge assessment of local evidence.  Appendix~\ref{app:training_config} documents the available training settings and gaps in the historical joint-run configuration.
Crucially, different tiles never share a GRPO group: doing so would turn differences in tile difficulty into artificial policy advantages.

%% file: sections/4_data.tex
\section{Data}
\label{sec:data}

\subsection{Main-Agent Trajectories}
The SFT corpus contains 2{,}197 main-agent trajectories and 4{,}267 sub-agent examples, with 6{,}336 training and 128 validation records.  After conversion and removal of RL-validation overlaps, the main-agent RL pool contains 1{,}997 prompts from ZwZ \citep{wei2026zoombench} and VisualProbe \citep{lai2026minio3}.  The 200-prompt RL validation set includes 196 earlier SFT training and four SFT validation examples, so external benchmarks provide the transfer evaluation.  Across nine source pools, exact and perceptual overlap exclusions change accuracy by at most $0.28$ points over 25 audited evaluations; neither check detects overlap on either HR-Bench split.  Appendix~\ref{app:data_audit} gives conversion counts, source composition, and overlap checks.

\subsection{Sub-Agent ROI Data}
Tile supervision is synthesized from VisualProbe and ZwZ-RL-VQA \citep{lai2026minio3,wei2026zoombench}.  ZwZ supplies absolute-pixel boxes; VisualProbe boxes are annotated by a strong vision-language model and then filtered.  Each source image yields a maximum-overlap candidate, an edge/partial-overlap candidate when available, and a zero-overlap candidate biased toward hard negatives.  Geometric candidate type is metadata rather than the final semantic label, since a zero-overlap tile may still contain duplicated or relational evidence.

A teacher model labels each tile using the question and an ROI hint, but the exported student prompt contains only the tile, question, tile ID, and tile coordinates.  A hint-free verifier rechecks suspicious positives, low-coverage answers, zero-overlap positives, and a random audit sample.  We then rewrite evidence strings that mention annotation artifacts or ROI hints, preserving the remaining structured fields.  The resulting training instance is a mapping from local pixels and a question to a structured evidence record.

\subsection{Difficulty and Mixture Control}
For RL, $K$ rollouts partition prompts into all-wrong, learnable, and all-correct groups.  Only mixed-reward groups have nonzero within-group GRPO advantage.  Some joint runs filter all-correct main prompts offline, and optional DAPO group filtering \citep{yu2025dapo} can remove zero-variance groups online; the main-only and forced-grid runs retain their full RL-validation-disjoint main pool without difficulty filtering.  The valid-grid warm start adds sub-agent examples to this pool.  Group filtering is distinct from handling a single capped trajectory.  Main and sub examples must be monitored separately because the pretrained sub task is substantially easier and optional group filtering can shift the effective main/sub mixture.

\subsection{Data Quality Limitations}
A human audit labeled 936 of 1{,}000 sampled examples: 491 from VisualProbe and 445 from ZwZ.  VisualProbe's model-produced boxes had median IoU $0.243$ with human boxes and were disjoint in $28.9\%$ of cases, while ZwZ boxes had median IoU $0.887$.  These source differences motivate verification and separate reporting of annotation quality.

%% file: sections/5_experiments.tex
\section{Experimental Setup}
\label{sec:experiments}

\subsection{Benchmarks}
We evaluate on TIR visual search (120 examples; \citealp{li2025tirbench}), VSTAR (191 gold-labeled examples; \citealp{wu2023vstar}), ZoomBench (845 examples; \citealp{wei2026zoombench}), and the HR-Bench 4K and 8K subsets (800 option-rotation rows each; \citealp{wang2025dc2}).  HR-Bench is also reported with CircularEval \citep{liu2024mmbench}, where a question is correct only if all four cyclic option rotations are answered correctly.  Main-text comparisons use these full denominators; Appendix~\ref{app:vstar_denominator} reports the VSTAR ambiguity-exclusion sensitivity.

\subsection{Models and Systems}
We test Qwen3.5 main agents at 4B and 9B \citep{qwen2026qwen35} and a Qwen3.6-27B main agent \citep{qwen2026qwen36}.  The default grid system uses a same-size Qwen sub-agent; selected settings use Gemini 3.5 Flash \citep{google2026gemini35flash} or a 27B Qwen sub-agent to examine the effect of tile-reader choice.  We compare two complete search policies:
\begin{itemize}
    \item \textbf{Zoom only}: a dedicated sequential-search policy using \texttt{zoom\_in}; and
    \item \textbf{Grid + zoom}: \method{} using \texttt{grid\_search} and \texttt{zoom\_in}.
\end{itemize}
We evaluate answer accuracy at each policy's inference settings; parallel tile inspection adds sub-agent computation.  Appendix~\ref{app:training_free_protocol} details the policy-specific prompts, inputs, iteration budgets, and replicate settings.

\subsection{Metrics and Statistical Protocol}
The primary metric is judge-all answer accuracy, with failed or unanswered gold examples counted as wrong.  Replicated cells report the mean and sample standard deviation over three runs unless marked otherwise, with per-arm seed conventions in Appendix~\ref{app:training_free_protocol}.  Post-training role comparisons use unseeded runs.  For paired comparisons, we evaluate both policies on the same examples.  A 95\% bootstrap confidence interval estimates uncertainty in their difference by repeatedly resampling those matched examples; an interval spanning zero leaves the direction unresolved at this level.  HR-Bench resampling keeps the four option rotations of each question together.  We also report tool calls, unfinished rate, visual-search hit rates when ROI labels exist, and CircularEval consistency.

\subsection{Post-Training Setup}
The SFT model is a LoRA adaptation \citep{hu2021lora} of Qwen3.5-4B and initializes the orchestration RL policies.  Main-only RL applies answer supervision to main-generated tokens; tool reports are observations, while shared weights can still change tile-reader behavior.  We select the policy on 200 RL-held-out prompts that largely appeared during SFT, using a greedy validation sweep.  Training uses eight prompts and eight rollouts per update, with the resolved optimizer, adapter, and sampling settings in Appendix~\ref{app:details}.

The first HR-Bench 4K role matrix crosses the SFT and main-only RL main agents with SFT and an earlier mixed-batch RL sub-agent, denoted early mixed RL.  It is distinct from the selected joint RL policy, which a separate matrix evaluates as main agent, sub-agent, and both.  That policy follows several warm-started joint-training stages rather than a single RL run initialized from SFT.

The HR-Bench 4K role comparisons use the same prompt, original-image crops, both search tools, a 15-iteration budget, and a Qwen3.6-27B judge on 800 option-rotation rows.  Failures count as wrong.  Both RL recipes sample $K=8$ responses per prompt, with distinct data, initialization, and adapter settings.  Joint training combines main-answer and direct tile-response losses (Eq.~\ref{eq:joint_objective}).  Zero-variance group filtering and training-time unfinished rewards are specified separately in the appendix.

%% file: sections/6_results.tex
\section{Results}
\label{sec:results}

\subsection{Parallel Search versus Sequential Zoom}
\begin{figure}[t]
\centering
\includegraphics[width=\linewidth]{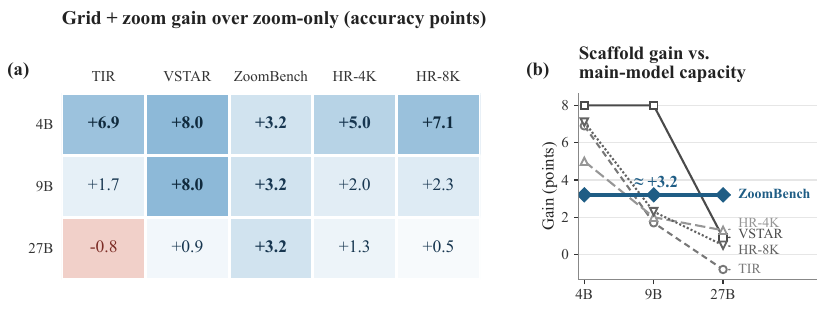}
\caption{Grid + zoom gain over dedicated zoom-only search.  (a) Differences in mean accuracy from Table~\ref{tab:grid_zoom}, in percentage points, with 14 of 15 cells positive.  (b) The same gains across model sizes; lines connect observations, without a fitted trend.  ZoomBench retains an approximately $+3.2$-point gain.  Colors encode differences, not statistical significance.  Per-arm replicate settings are in Appendix~\ref{app:training_free_protocol}.}
\label{fig:grid_zoom_gain}
\end{figure}

Table~\ref{tab:grid_zoom} summarizes the central training-free comparison, and Figure~\ref{fig:grid_zoom_gain} visualizes its gains.  \method{} improves mean accuracy over dedicated zoom-only search in 14 of 15 same-model benchmark cells.  Gains are largest for smaller main agents on TIR, VSTAR, and HR-Bench, while ZoomBench shows an approximately $3.2$ point gain at every tested model size.  At 27B, only the ZoomBench ($+3.2$) and HR-Bench 4K ($+1.3$) gaps exceed the replicate spread of both arms.  The remaining 27B gaps, including the $-0.8$ point TIR difference, lie inside replicate variation.

\begin{table*}[t]
\centering
\caption{Judge-all accuracy (\%) for the two training-free search policies (Appendix~\ref{app:training_free_protocol}). Grid + zoom uses a same-size tile reader; $\dagger$ replaces it with Gemini 3.5 Flash. Values are mean $\pm$ sample standard deviation over three runs; $\Delta$ compares the same-model grid and zoom-only arms using unrounded means. Shading marks same-model grid search; bold marks the higher mean in each same-model pair; blue/orange deltas mark increases/decreases, not significance. }

\label{tab:grid_zoom}
\small
\setlength{\tabcolsep}{3pt}
\begin{tabular}{llrrrrr}
\toprule
Main & Recipe & TIR & VSTAR & ZoomBench & HR-4K & HR-8K \\
\midrule
\rowcolor{tablefocus}
27B & Grid + zoom & $76.1\pm0.96$ & \best{$94.4\pm1.60$} & \best{$67.8\pm0.94$} & \best{$90.1\pm0.50$} & \best{$89.3\pm0.59$} \\
    & Grid + zoom$^{\dagger}$ & $77.8\pm2.93$ & $96.7\pm1.32$ & $67.9\pm0.30$ & $90.4\pm0.07$ & $90.9\pm0.40$ \\
    & Zoom only   & \best{$76.9\pm0.48$} & $93.5\pm0.60$ & $64.6\pm1.30$ & $88.8\pm0.19$ & $88.8\pm0.40$ \\
    & $\Delta$    & \negdelta{-0.8} & \posdelta{+0.9} & \posdelta{+3.2} & \posdelta{+1.3} & \posdelta{+0.5} \\
\midrule
\rowcolor{tablefocus}
9B  & Grid + zoom & \best{$66.4\pm2.93$} & \best{$85.3\pm2.28$} & \best{$58.4\pm1.54$} & \best{$84.0\pm0.80$} & \best{$80.5\pm0.75$} \\
    & Zoom only   & $64.7\pm1.92$ & $77.3\pm5.45$ & $55.2\pm1.66$ & $82.0\pm0.47$ & $78.3\pm1.45$ \\
    & $\Delta$    & \posdelta{+1.7} & \posdelta{+8.0} & \posdelta{+3.2} & \posdelta{+2.0} & \posdelta{+2.3} \\
\midrule
\rowcolor{tablefocus}
4B  & Grid + zoom & \best{$59.4\pm4.59$} & \best{$78.4\pm2.47$} & \best{$51.0\pm1.03$} & \best{$78.6\pm1.08$} & \best{$76.1\pm0.22$} \\
    & Zoom only   & $52.5\pm1.44$ & $70.3\pm1.60$ & $47.8\pm1.30$ & $73.6\pm0.36$ & $69.0\pm0.33$ \\
    & $\Delta$    & \posdelta{+6.9} & \posdelta{+8.0} & \posdelta{+3.2} & \posdelta{+5.0} & \posdelta{+7.1} \\
\bottomrule
\end{tabular}
\end{table*}

\subsection{Where the Scaffold Helps}
The advantage of \method{} is generally larger at the smaller tested model sizes.  On HR-4K/8K it falls from $+5.0/+7.1$ points at 4B to $+1.3/+0.5$ at 27B.  With the 4B main agent fixed, replacing its tile reader with a 27B model adds $+1.1/+2.0$ points on HR-4K/8K.  These results suggest that controller-side evidence acquisition and aggregation remain a substantial bottleneck in the tested systems, even with a stronger local reader.

With the 27B main agent, Gemini 3.5 Flash tile readers improve the mean by about $1.7$ points on TIR and $2.3$ points on VSTAR, with smaller gains on the other benchmarks (Table~\ref{tab:grid_zoom}).

CircularEval tests whether the advantage persists when success requires correct answers under all four option rotations (Table~\ref{tab:hr_circular}).  \method{} has the higher mean in all six model--resolution pairs.  At 4B, HR-Bench 4K/8K consistency rises from $48.2/40.2\%$ with dedicated zoom-only search to $59.0/52.8\%$ with grid + zoom.  The smaller gaps at 27B follow the pattern in row-level accuracy.  Thus, the observed benefit extends to questions answered consistently across option orderings, beyond isolated correct rows.

The absolute consistency gap remains substantial.  For the same-model grid arms, HR-Bench 4K/8K row accuracy of $78.6/76.1\%$ at 4B and $84.0/80.5\%$ at 9B falls to CircularEval $59.0/52.8\%$ and $68.3/62.0\%$.  Better evidence acquisition and answer consistency are therefore complementary targets: the search scaffold improves the stricter metric, while small models remain sensitive to option ordering.

\begin{table}[t]
\centering
\caption{HR-Bench CircularEval means (\%), computed over 200 questions per subset.  A question counts as correct only when all four option rotations are correct.  These values are separate from the 800-row accuracy in Table~\ref{tab:grid_zoom}. Shading marks grid search; bold marks the higher mean within each model size.}
\label{tab:hr_circular}
\small
\begin{tabular}{llcc}
\toprule
Main & Search & 4K & 8K \\
\midrule
\rowcolor{tablefocus}
4B & Grid + zoom & \best{59.0} & \best{52.8} \\
4B & Dedicated zoom & 48.2 & 40.2 \\
\rowcolor{tablefocus}
9B & Grid + zoom & \best{68.3} & \best{62.0} \\
9B & Dedicated zoom & 65.3 & 57.2 \\
\rowcolor{tablefocus}
27B & Grid + zoom & \best{81.2} & \best{79.3} \\
27B & Dedicated zoom & 80.8 & 78.8 \\
\bottomrule
\end{tabular}
\end{table}

\begin{table*}[t]
\centering
\caption{Pretrained, SFT, and RL policies (accuracy, \%). (a) Descriptive cross-benchmark results: three-run pretrained means and single post-training runs. (b) Main-only / early mixed RL role matrix, first of three passes. (c) Selected joint RL role matrix, one pass with its own SFT baseline. VSTAR uses all 191 gold examples; failures count as wrong. Appendix~\ref{app:sft_comparison} gives per-run counts and evaluation settings. Shading identifies SFT references; bold marks the highest observed accuracy per column within each panel, not significance.}
\label{tab:post_training}
\small
\setlength{\tabcolsep}{3pt}
\begin{tabular}{llrrrrr}
\toprule
\multicolumn{7}{l}{\textbf{(a) Cross-benchmark evaluations}} \\
Main policy & Sub policy & TIR & VSTAR & ZoomBench & HR-4K & HR-8K \\
\midrule
Pretrained & Pretrained & 59.44 & 78.36 & 51.01 & 78.58 & 76.12 \\
\rowcolor{tablefocus}
SFT & SFT & 72.50 & \best{87.96} & 60.00 & \best{82.75} & 82.25 \\
Main-only RL & SFT & \best{74.17} & 86.91 & \best{61.66} & \textit{see (b)} & \best{82.75} \\
\bottomrule
\end{tabular}

\vspace{5pt}
\begin{minipage}[t]{0.49\linewidth}
\centering
\small
\begin{tabular}{lrr}
\toprule
\multicolumn{3}{l}{\textbf{(b) Main-only / early mixed RL: HR-4K}} \\
Main policy & SFT sub & Early mixed sub \\
\midrule
\rowcolor{tablefocus}
SFT & 84.88 & 83.38 \\
Main-only RL & \best{85.75} & \best{85.75} \\
\bottomrule
\end{tabular}
\end{minipage}\hfill
\begin{minipage}[t]{0.49\linewidth}
\centering
\small
\begin{tabular}{lrr}
\toprule
\multicolumn{3}{l}{\textbf{(c) Selected joint RL: HR-4K}} \\
Main policy & SFT sub & Joint RL sub \\
\midrule
\rowcolor{tablefocus}
SFT & \best{84.50} & \best{85.88} \\
Joint RL & 82.12 & 82.00 \\
\bottomrule
\end{tabular}
\end{minipage}
\end{table*}

\subsection{Supervised Fine-Tuning}
SFT raises observed accuracy across all five evaluated benchmark splits (Table~\ref{tab:post_training}): TIR reaches $72.50\%$, VSTAR $87.96\%$, ZoomBench $60.00\%$, and HR-Bench 4K/8K $82.75/82.25\%$.  The matched unseeded HR-Bench 4K comparison gives a $4.17$-point gain with a 95\% question-paired bootstrap interval of $[+1.54,+6.75]$ points, conditional on the evaluated runs.  The other splits provide descriptive comparisons; Appendix~\ref{app:sft_comparison} gives their decoding settings and run counts.

\subsection{Main-Only RL and Tool Use}
Main-only RL primarily adapts tool use.  In an internal 100-prompt evaluation with four responses per prompt and the same policy in both roles, mean tool calls fall from $2.65$ to $2.11$ (paired difference $-0.54$, 95\% interval $[-0.87,-0.21]$).  Pass@1 is $71.25\%$ versus $72.50\%$, with an interval spanning zero for the difference, while the pass@4 point estimate falls from $89\%$ to $83\%$.  Pass@4 counts a prompt as solved if any of its four responses is correct.  This evaluation supports fewer calls per trajectory; total token cost and latency were not recorded.  The shared policy changes both deployed roles, so the comparison does not isolate the controller.

With the SFT tile reader fixed, HR-Bench 4K accuracy rises from $84.88\%$ to $85.75\%$; the $0.88$-point difference has a 95\% question-paired interval of $[-1.38,+3.25]$ points.  The second pass gives the same difference, again with an interval spanning zero.  The third early mixed RL pass preserves the positive main-agent direction, without a reported paired interval.  Cross-benchmark results are mixed (Table~\ref{tab:post_training}a): TIR and ZoomBench increase, VSTAR decreases, and HR-Bench 8K changes from $82.25\%$ to $82.75\%$.  These cross-benchmark point estimates are descriptive (Appendix~\ref{app:sft_comparison}).  An earlier VSTAR comparison deploying each policy in both roles gives $88.0\%$ for each.  Together, the evaluations support search-behavior adaptation without establishing a replicated external accuracy gain.

\subsection{Joint Training Dynamics and Role Asymmetry}
Our paired GRPO recipe trains both roles through separate main-answer and local-evidence objectives (Figure~\ref{fig:paired_grpo}).  This preserves prompt-matched relative comparisons while coupling the two roles through shared model weights.

During joint training, aggregate pass@1 rises while sampled pass@8 remains flat.  This is consistent with reinforcing successful responses, although the changing training prompts prevent a fixed-set coverage comparison.

In a single unseeded deployment-role matrix, the two roles respond differently to this training.  With the SFT main agent, the joint RL tile reader raises HR-Bench 4K accuracy from $84.50\%$ to $85.88\%$, a $1.38$-point difference with a 95\% interval of $[-1.12,+3.88]$ points.  Deploying joint RL as the main agent instead lowers accuracy and completion (Table~\ref{tab:post_training}).  With the joint tile reader fixed, its $3.88$-point accuracy drop has a question-level interval of $[-6.88,-1.12]$ points.  This separates local-evidence adaptation from the main policy's ability to finish a search.

Joint RL as main completes 68--69 fewer of the 800 rows, with additional failures concentrated at iteration and context limits.  Its mean tool calls rise from $2.22$ to $4.35$ with the SFT sub-agent fixed.

%% file: sections/8_conclusion.tex
\section{Conclusion}
\label{sec:conclusion}

\method{} combines parallel, question-conditioned tile inspection with adaptive zoom to make high-resolution evidence acquisition both effective and trainable.  Across five benchmark splits and three model sizes, the deployed scaffold improves mean accuracy over dedicated zoom-only search in 14 of 15 cells, with larger gains for smaller main agents.  SFT raises observed accuracy across all five splits, including a $4.17$-point HR-Bench 4K gain.

The paired role-specific GRPO surrogate gives the controller and tile reader distinct learning signals and makes the omitted cross-role credit paths explicit.  Main-only RL reduces tool use while maintaining similar pass@1 in the internal four-response evaluation.  Joint training reveals a complementary optimization challenge: local-reader adaptation and successful global search do not move together, and deploying the joint-trained controller reduces completion.  Together, these findings support \method{} as a trainable decomposition for visual search and identify efficient evidence acquisition and reliable termination as central targets for controller optimization.

%% file: sections/2_related_work.tex
\section{Related Work}
\label{sec:related_work}

\paragraph{Learned agent orchestration.}
Language-agent systems have progressed from optimizing fixed computational graphs and code-represented workflows \citep{zhuge2024gptswarm,hu2025adas,zhang2025aflow} to selecting a query-dependent architecture \citep{zhang2025maas} or dynamically activating agents with a learned controller \citep{dang2025puppeteer}.  Recent work makes the orchestrator itself the primary learning target.  ToolOrchestra trains a compact controller to route among models and tools with outcome, efficiency, and preference rewards \citep{su2025toolorchestra}; AOrchestra creates a task-specific executor by choosing its instruction, context, tools, and model \citep{ruan2026aorchestra}; and MAS-Orchestra uses function-calling RL to construct a complete multi-agent system, accompanied by MASBench axes that explicitly include task breadth and parallelism \citep{ke2026masorchestra}.  Closest to our execution pattern, ParaManager emits parallel batches over a unified agent/tool interface \citep{yuan2026paramanager}, while Orchestra-o1 performs modality-aware decomposition, parallel specialist execution, and decision-level RL for an omnimodal main agent \citep{zhang2026orchestrao1}.

\method{} adopts this orchestrator--worker abstraction in a deliberately constrained domain.  Unlike methods that search arbitrary roles, models, tools, or communication graphs, our workers are homogeneous tile readers and decomposition is geometrically fixed; the learned decisions are when to fan out, how to interpret and aggregate structured local evidence, where to inspect next, and when to stop.  This restriction turns high-resolution image search into a controlled orchestration testbed: spatial tiles provide auditable subproblems, parallelism is real rather than simulated, and worker outputs can be checked against region annotations.  It also distinguishes our training setup from work that trains only the controller: we construct separate, prompt-matched groups and rewards for both the main and sub-agent roles, while treating their shared-parameter combination as a multi-task surrogate rather than exact end-to-end credit assignment.

\paragraph{Active visual evidence acquisition.}
High-resolution visual reasoning exposes a coverage--resolution trade-off.  V* introduced LLM-guided visual search and V*Bench \citep{wu2023vstar}; DC$^2$ recursively partitions images and converts local patches into textual evidence, alongside HR-Bench \citep{wang2025dc2}; and ZoomEye performs hierarchical tree search over image regions \citep{shen2025zoomeye}.  TIR-Bench broadens evaluation to tool-mediated image transformations, of which we use its visual-search subset \citep{li2025tirbench}, whereas ZoomBench measures the gap between global and cropped fine-grained perception \citep{wei2026zoombench}.  Mini-o3 scales multi-turn visual search and releases VisualProbe, which supplies part of our tile-supervision images \citep{lai2026minio3}.  These works establish the visual task, but our central comparison is orchestration: dedicated sequential \texttt{zoom\_in} asks one agent to propose and inspect every region, while \method{} first fans out question-conditioned local reading and then lets the main agent merge evidence and choose a precise crop.  WideSearch studies the analogous completeness problem for broad web information seeking \citep{wong2025widesearch}, reinforcing that coverage and aggregation---not only difficult local reasoning---can bottleneck agents.

\paragraph{Post-training orchestrators and workers.}
We use supervised trajectory distillation and parameter-efficient adaptation \citep{hinton2015distilling,hu2021lora}, followed by a GRPO-style objective \citep{shao2024deepseekmath}.  In contrast to outcome-only controller training, Orchestra-o1 supplies offline, decision-level rubric rewards \citep{zhang2026orchestrao1}, ParaManager combines task, protocol, diversity, and efficiency rewards over orchestration trajectories \citep{yuan2026paramanager}, and LEMON applies localized counterfactual reward contrasts to edited fields of an orchestration specification \citep{chen2026lemon}.  Our setting adds a second conditional distribution: local-reader responses have dense geometric and schema rewards, while main-agent trajectories receive end-task reward.  We therefore normalize only among responses to an identical role-specific prompt and combine the two role losses explicitly; mixing different tiles in one relative-reward group would incorrectly treat subproblem difficulty as policy quality.  This construction supplies direct local supervision but deliberately omits the high-variance global-return score term on nested tile-reader samples.  Recent evidence that tool-call text can account for gains attributed to returned crops \citep{shao2026textcall} identifies a useful future control: separating the contribution of the returned visual evidence from that of the tool-call text.

%% file: sections/7_analysis.tex
\section{Analysis and Limitations}
\label{sec:analysis}

\paragraph{Search Scaffolds and Main-Agent Models}
The pattern across tested models in Figure~\ref{fig:grid_zoom_gain} suggests that parallel tile inspection supplies an explicit coverage strategy that smaller main agents do not reliably implement by themselves.  A stronger tile reader helps only modestly when the main agent is weak, whereas the 27B main agent shows a smaller grid-versus-zoom gap on most benchmarks.  These operating points span Qwen3.5 and Qwen3.6 and do not isolate the effect of parameter count.  ZoomBench is the notable exception: its persistent grid advantage is consistent with tasks that reward broad spatial scanning.

\paragraph{Trajectory Variance Is Part of the Task}
Repeated rollouts of the same checkpoint and prompt frequently disagree even when token decoding is nominally deterministic.  A small difference in the first region changes the crop, subsequent evidence, and final answer.  Single-rollout validation therefore confounds policy quality with path variance.  We recommend prompt-paired, multi-rollout evaluation and reporting both mean accuracy and within-prompt dispersion.

\paragraph{RL Consolidates More Readily Than It Discovers}
Joint RL makes this pattern quantitative.  Training pass@1 rises while sampled pass@8 stays flat.  Because these aggregates use changing training prompts, they are consistent with concentration on successes but do not measure coverage change on a fixed set.  The same reading fits the held-out main-only checks: accuracy point estimates are small, the high-coverage pass@4 point estimate falls, and the resolved change is fewer tool calls.  An earlier small-pool main-only RL run shows the item-level version of the gap: held-out answer flips are nearly symmetric, and overall accuracy does not rise.  Sparse end-task reward can reinforce an available search path; it has not yet provided a reliable signal for discovering a new region-selection strategy.

The external joint RL role matrix also separates the two deployment roles.  Replacing only the SFT tile reader by joint RL gives an unresolved $+1.38$ point estimate.  Deploying joint RL as the main agent lowers accuracy by $2.38$ points with the SFT sub-agent and $3.88$ points with the joint RL sub-agent; only the latter fixed-sub-agent contrast has a paired test excluding zero.  Completion falls from $789$ to $720$ rows with the SFT sub-agent and from $790$ to $722$ with the joint RL sub-agent.  Most additional failures reach the iteration or context limit.  The result locates a deployment bottleneck in the main agent's evidence acquisition and completion, despite the positive sub-agent point estimate.

The exploratory grid interventions probe whether a policy can use fixed tile evidence and whether grid-call preference can be increased.  Under a forced single grid and no zoom, the selected policy reached $65.5\%$ on the 200-prompt internal validation set; when restored to free grid-plus-zoom use, it reached $67.5\%$.  The subsequent valid-grid warm start raised successful grid use from $9.5\%$ to $21.0\%$, while accuracy across its saved policies declined from $71.5\%$ to $64.5\%$ and unfinished responses rose to $10.0\%$.  Its early $71.5\%$ point estimate versus the $67.5\%$ warm-start baseline has a paired interval spanning zero.  The changed loop, mixed-role data, and missing synchronous no-shaping control prevent attribution of these trends to the reward term alone.  The validation set is RL-held-out but SFT-exposed, so this is a mechanism observation rather than an external accuracy gain.

\paragraph{Local Supervision and Main-Agent Credit Differ}
Sub-agent ROI rewards are dense and interpretable, but they do not directly assign credit to the main agent's decision to call grid search or select a region.  If both roles are treated as trainable policies, the exact global-return gradient also contains a score term for each sampled sub-agent report.  We omit this term because terminal reward broadcast across many reports has poor credit specificity, and use fixed external prompts for direct local supervision.  Such prompts have no main-policy occupancy derivative; a grid-derived replay alternative would omit that derivative by freezing the sampled prompts for each update.  Consequently, paired training with shared weights is a biased, lower-variance multi-task semi-gradient, not causal credit assignment across the main/sub boundary.  Moreover, masking a role's response tokens does not freeze that role when the backbone is shared: either role loss can still change both deployed policies.  Future work should test frozen or separate role adapters, downstream evidence attribution, learned search-value models, or context-matched counterfactual interventions.

\paragraph{Engineering Choices Can Masquerade as Algorithmic Results}
The development runs uncovered several high-impact confounders: thumbnail rather than original-image cropping, truncated grid reports, mismatched tool descriptions, missing configuration propagation, optional zero-variance group filtering, and evaluation prompts with incompatible sampling settings.  In particular, one apparent null effect of \texttt{grid\_search} was later invalidated because $97.7\%$ of grid reports were truncated.  Capped trajectories introduce a separate ambiguity: they normally remain in the batch, and the recorded \texttt{unfinished\_reason} does not universally force terminal correctness to zero.  Observation completeness, image provenance, group-filter configuration, and unfinished-reward semantics therefore bound the interpretation of these diagnostic runs.

The matched unfinished-reward pilot now supplies a controlled intervention on this termination rule.  Its final internal evaluation favors explicit failure penalization in both accuracy and completion, while intermediate effects vary (Appendix~\ref{app:hardfail}).  The SFT-exposed evaluation and changing intermediate effects limit this result to an internal diagnostic.

\paragraph{Limitations}
Our benchmark suite is dominated by high-resolution VQA and does not yet isolate counting, relational, OCR, and document-layout failures with equal coverage.  Some replicate protocols differ across model sizes, the 9B service is no longer reproducible in its original environment, and two HR-Bench control arms have only two repetitions.  The joint RL role matrix is a single unseeded pass; its question-level bootstrap does not measure variability across repeated executions.  ROI labels include model-generated and noisy boxes.  The internal RL validation set is SFT-exposed.  The nine-pool audit finds exact byte and decoded-RGB overlaps only on ZoomBench, plus perceptual candidates on ZoomBench and TIR.  Lineage-specific exclusions change reported accuracy by at most $0.28$ points; no overlap is detected on VSTAR or either HR-Bench split under these checks (Appendix~\ref{app:data_audit}).  Perceptual checks remain a sensitivity analysis rather than proof of contamination or its absence.  Finally, our current implementation performs only one grid layer; recursive or asynchronous search may improve recall but would also complicate context management and credit assignment.

%% file: sections/9_appendix.tex
\section{Additional Experimental Details}
\label{app:details}

\subsection{Training-Free Replicate Protocols}
\label{app:training_free_protocol}
Table~\ref{tab:grid_zoom} and Figure~\ref{fig:grid_zoom_gain} use three runs per arm and a common Qwen3.6-27B judge.  On TIR, VSTAR, and ZoomBench, all grid arms and the 27B dedicated zoom-only arm use explicit seeds; the 4B dedicated zoom-only arm uses unseeded repetitions, while the 9B dedicated zoom-only arm combines an unseeded first run with two explicitly seeded runs.  All HR-Bench repetitions are unseeded.  Standard deviations in Table~\ref{tab:grid_zoom} are sample standard deviations, and differences use unrounded means.  The two search policies use different tool prompts: grid + zoom uses a grid-and-zoom prompt, up to 15 main iterations, and image dimensions supplied in the input; dedicated zoom-only uses a zoom-only prompt, up to 20 iterations, and no dimension injection.  Each grid call additionally dispatches $n^2$ tile-reader requests.  These cells measure the complete search policies at their respective inference settings, without isolating the grid operator or matching total inference compute.

Table~\ref{tab:training_free_runs} gives the archived numerators for the self-sub-agent grid arm and its dedicated zoom-only comparator.  Each slash-separated triplet lists the three run numerators in run order, with one common denominator per benchmark.  Gold examples that failed to produce an answer remain in the denominator.  The seeded 27B zoom-only run counts reproduce the means and sample deviations in Table~\ref{tab:grid_zoom}.  The earlier single unseeded 27B zoom-only run ($94/120$, $174/191$, $552/845$) is excluded from that table.  The 9B zoom-only run labeled first in its archived campaign is an unseeded baseline, followed by runs with explicit seeds; it is grouped with that campaign in the table, without claiming all three draws used the same seed convention.

\begin{table*}[t]
\centering
\caption{Training-free run numerators.  Within each cell the three counts share the denominator in the column heading.  Grid runs on these three benchmarks are seeded; the 4B dedicated zoom-only runs are unseeded.  The 9B zoom-only first run is an unseeded baseline followed by two seeded runs. Shading identifies the grid-search reference rows.}
\label{tab:training_free_runs}
\small
\setlength{\tabcolsep}{4pt}
\begin{tabular}{lllccc}
\toprule
Main & Search & Replicates & TIR /120 & VSTAR /191 & ZoomBench /845 \\
\midrule
\rowcolor{tablefocus}
4B & Grid + zoom & seeds 1--3 & 74/75/65 & 146/155/148 & 427/441/425 \\
4B & Dedicated zoom & reps 1--3 & 62/65/62 & 137/131/135 & 411/391/409 \\
\rowcolor{tablefocus}
9B & Grid + zoom & seeds 1--3 & 83/80/76 & 161/160/168 & 493/481/507 \\
9B & Dedicated zoom & base, seeds 2--3 & 79/75/79 & 151/156/136 & 462/455/482 \\
\rowcolor{tablefocus}
27B & Grid + zoom & seeds 1--3 & 90/92/92 & 183/181/177 & 582/570/567 \\
27B & Dedicated zoom & seeds 1--3 & 92/93/92 & 180/178/178 & 552/552/533 \\
\bottomrule
\end{tabular}
\end{table*}

The corresponding failed-trajectory counts for the 4B grid runs are $12/6/17$ on TIR, $24/13/18$ on VSTAR, and $60/53/70$ on ZoomBench; for 4B zoom-only they are $9/18/17$, $22/27/35$, and $69/78/71$.  For 9B grid the counts are $5/11/5$, $12/7/7$, and $8/16/5$; for 9B zoom-only they are $0/7/6$, $1/15/18$, and $2/19/14$.  The 4B-grid VSTAR counts are gold rows not completed in the final scored outputs; the other counts retain the historical run-status definition. These are not additional errors to subtract from accuracy numerators.  In particular, the 9B zoom-only third VSTAR run has 18 failures and $136/191$ correct, explaining much of that arm's spread.

For 27B self-sub-agent grid, the failed counts are $2/4/1$ on TIR, $2/2/1$ on VSTAR, and $1/5/0$ on ZoomBench.  For seeded 27B zoom-only, the counts of rows without completion in the reconciled run summaries are $4/2/1$, $4/1/0$, and $8/8/6$, respectively.

\begin{table}[t]
\centering
\caption{HR-Bench run numerators for the same training-free arms.  All repetitions are unseeded and each subset has 800 option-rotation rows. Shading identifies the grid-search reference rows.}
\label{tab:hr_runs}
\small
\begin{tabular}{llcc}
\toprule
Main & Search & 4K /800 & 8K /800 \\
\midrule
\rowcolor{tablefocus}
4B & Grid + zoom & 627/638/621 & 610/607/610 \\
4B & Dedicated zoom & 592/587/587 & 553/549/554 \\
\rowcolor{tablefocus}
9B & Grid + zoom & 679/669/667 & 650/638/645 \\
9B & Dedicated zoom & 653/654/660 & 614/628/637 \\
\rowcolor{tablefocus}
27B & Grid + zoom & 717/725/721 & 716/709/718 \\
27B & Dedicated zoom & 709/710/712 & 713/712/707 \\
\bottomrule
\end{tabular}
\end{table}

The failed-trajectory triplets corresponding to the six HR-Bench rows in Table~\ref{tab:hr_runs}, in the same order, are $24/25/30$ and $39/50/39$ (4B grid, 4K/8K), $51/60/44$ and $71/85/69$ (4B dedicated zoom), $7/11/5$ and $20/22/23$ (9B grid), $7/11/8$ and $31/17/16$ (9B dedicated zoom), $8/3/6$ and $18/16/12$ (27B grid), and $3/8/8$ and $15/17/12$ (27B dedicated zoom).  HR-Bench CircularEval in Table~\ref{tab:hr_circular} measures a stricter 200-question event and cannot be obtained by dividing these 800-row counts by four.

\subsection{Gemini Tile Readers and Direct Answers}
\label{app:gemini}
Table~\ref{tab:gemini_tile} gives the run counts underlying the replicated Gemini row in Table~\ref{tab:grid_zoom} and the earlier single-run prompt variants.  All displayed values use the Qwen3.6-27B judge-all evaluator.  The V1 entries are single runs; the V2 27B entries are three seeded runs.  The V1 and V2 prompts differ, so the two sets describe distinct operating points.  Under V2, Gemini 3.5 Flash raises the 27B mean by $1.7$ and $2.3$ points on TIR and VSTAR relative to self-sub-agent grid search, while ZoomBench changes by about $0.1$ point.  The HR-Bench Gemini arm also uses three unseeded runs: its 4K numerators are $724/723/723$ and its 8K numerators are $726/731/725$, each out of 800.  Its CircularEval means are $82.3\%$ and $81.7\%$, respectively.

\begin{table*}[t]
\centering
\caption{Gemini 3.5 Flash as the grid tile reader.  V1 entries are single-run correct/total values; V2 27B entries show three seeded numerators with the denominator in the heading.  All cells use judge-all scoring and count failed gold examples as wrong. Shading identifies the replicated V2 setting; V1 and V2 are not matched comparisons.}
\label{tab:gemini_tile}
\small
\begin{tabular}{llccc}
\toprule
Main & Prompt & TIR /120 & VSTAR /191 & ZoomBench /845 \\
\midrule
4B & V1 & 84 & 168 & 527 \\
9B & V1 & 89 & 166 & 496 \\
27B & V1 & 97 & 188 & 599 \\
\rowcolor{tablefocus}
27B & V2, seeds 1--3 & 93/97/90 & 187/185/182 & 574/571/576 \\
\bottomrule
\end{tabular}
\end{table*}

No-search TIR controls are excluded from the quantitative comparison because the available summaries disagree on their Qwen and Gemini numerators and do not identify a reconciled scored run.  The historical V1 four-benchmark report also includes VTC, but its original aggregate used rule scoring; VTC is omitted rather than mixed with judge-all cells.

\subsection{Post-Training Checkpoint Ledger}
\begin{table}[t]
\centering
\caption{Three unseeded HR-Bench 4K passes of the early mixed RL role matrix. Correct counts share an 800-row denominator; failures count as wrong. All use V2 prompts, original-image crops, and Qwen3.6-27B judge-all. The last column gives completed rows in pass 3. The mixed reader is from the early mixed RL lineage, not the selected joint policy. Shading marks SFT main rows; Table~\ref{tab:post_training}(b) retains pass 1.}
\label{tab:role_matrix}
\small
\setlength{\tabcolsep}{4pt}
\begin{tabular}{llrrrr}
\toprule
& & \multicolumn{3}{c}{Correct /800} & Complete /800 \\
\cmidrule(lr){3-5}
Main & Sub & Pass 1 & Pass 2 & Pass 3 & Pass 3 \\
\midrule
\rowcolor{tablefocus}
SFT & SFT & 679 & 663 & 659 & 786 \\
\rowcolor{tablefocus}
SFT & Early mixed & 667 & 667 & 680 & 788 \\
Main-only RL & SFT & 686 & 670 & 688 & 785 \\
Main-only RL & Early mixed & 686 & 682 & 687 & 784 \\
\bottomrule
\end{tabular}
\end{table}

\begin{table*}[t]
\centering
\caption{HR-Bench 4K deployment-role matrix for the selected joint RL policy.  Each cell is a single unseeded run reporting judge-all accuracy and completed trajectories over the same 800 option-rotation rows; failures count as wrong. Shading marks the SFT main; bold marks the higher accuracy within each sub-agent setting, not significance.}
\label{tab:joint_v4_role_matrix}
\small
\begin{tabular}{lrrrr}
\toprule
& \multicolumn{2}{c}{SFT sub} & \multicolumn{2}{c}{Joint RL sub} \\
\cmidrule(lr){2-3}\cmidrule(lr){4-5}
Main & Accuracy & Complete & Accuracy & Complete \\
\midrule
\rowcolor{tablefocus}
SFT & \best{84.50} (676/800) & 789 & \best{85.88} (687/800) & 790 \\
Joint RL & 82.12 (657/800) & 720 & 82.00 (656/800) & 722 \\
\bottomrule
\end{tabular}
\end{table*}
Table~\ref{tab:role_matrix} combines the SFT cold start, a main-only RL policy initialized from it, and an early mixed RL sub-agent.  The independently evaluated joint RL policy in Table~\ref{tab:joint_v4_role_matrix} follows a separate, later joint-training lineage.  Its preceding joint-training stage is distinct from the early mixed RL sub-agent.  The role and lineage labels therefore identify different policies even when they share an SFT ancestor.

The formal role-matrix cells are the deduplicated judge outputs of the first HR-Bench 4K pass.  The second pass scores $82.88/83.38/83.75/85.25\%$ for SFT/SFT, SFT/early mixed RL sub-agent, main-only RL/SFT, and main-only RL/early mixed RL sub-agent.  The formal numerators are $679/800$, $667/800$, $686/800$, and $686/800$; the repetition is a separate pass.  The third pass scores $82.38/85.00/86.00/85.88\%$ in the same role order, with $786/788/785/784$ completed rows (Table~\ref{tab:role_matrix}).  It repeats the early mixed RL matrix only; the selected joint RL matrix remains a single pass.

\subsection{VSTAR Denominator Reconciliation}
\label{app:vstar_denominator}
The earlier shared-actor evaluation deploys each policy in both roles; its SFT and main-only RL runs contain the same 191 gold-labeled VSTAR test rows.  The archived evaluator excludes two rows as ambiguous: \texttt{sa\_25678}, a viewpoint-dependent left/right question, and \texttt{sa\_40188}.  Thus the reported denominator of 189 is an evaluation exclusion, not missing gold or missing predictions.  Restoring both rows gives $168/191$ ($88.0\%$) for each policy, with 11 RL-only successes, 11 SFT-only successes, and 169 ties.  Applying the archived ambiguity filter gives $167/189$ ($88.4\%$) for each, with 10 wins, 10 losses, and 169 ties.  SFT completes all 189 retained rows; RL completes 187 and its two iteration-limit failures count as wrong.  Each policy answers one of the excluded rows correctly, but they succeed on opposite rows.  The zero net difference therefore holds under both denominator conventions.

\subsection{Earlier Main-Only RL Checkpoints}
\label{app:earlier_rl}
Earlier main-only checkpoints agree with this picture.  With a fixed sub-agent, an RL policy trained on 64 prompts is $-0.53$ points on VSTAR and $-0.83$ points on TIR relative to SFT.  On 64 training items its pass@1 rose, but 50 of 309 held-out items flipped: SFT alone was correct on 26 and RL alone was correct on 24.  Another earlier main-only policy with the SFT sub-agent is below the paired SFT baseline on both HR-Bench 4K ($81.69$ vs.\ $82.97$) and VSTAR ($87.30$ vs.\ $88.36$).  Pairing that same checkpoint with its own sub-agent gives a $+1.41$ point HR-Bench 4K estimate whose McNemar test \citep{mcnemar1947note} is $p=0.32$, while VSTAR remains below the SFT baseline.

\subsection{SFT Comparison and Uncertainty}
\label{app:sft_comparison}
Table~\ref{tab:post_training_counts} gives the single-run counts behind the cross-benchmark block of Table~\ref{tab:post_training}.  Both policies use the V2 grid-and-zoom harness, original images with dimensions, a 15-iteration budget, unseeded decoding, and the same judge family.  The later main-only RL evaluation fixes the tile reader to SFT.  Serving endpoints and runtime versions differ between the historical SFT and later RL evaluations.  HR-Bench 8K uses the clean SFT/SFT evaluation.  These cross-benchmark SFT/RL comparisons remain descriptive, without paired uncertainty estimates.  The historical pretrained references on TIR, VSTAR, and ZoomBench additionally use explicit seeds.  Their SFT differences are $+13.06$, $+9.60$, and $+8.99$ points, respectively.  The formal HR-Bench 4K role comparisons retain their separate SFT baselines in Table~\ref{tab:role_matrix}.

\begin{table}[t]
\centering
\caption{Single-run SFT and later main-only RL results with the SFT tile reader fixed. Entries give correct and completed counts; all gold examples remain in the accuracy denominator. The HR-4K SFT row is the historical comparison, not the later role-matrix baseline; HR-8K uses the clean SFT/SFT replacement. Shading identifies benchmarks with both runs. Dashes denote unavailable same-campaign cells; the separate HR-4K role matrix is in Table~\ref{tab:role_matrix}.}
\label{tab:post_training_counts}
\small
\begin{tabular}{lrrrrr}
\toprule
& Gold & \multicolumn{2}{c}{SFT main} & \multicolumn{2}{c}{RL main} \\
\cmidrule(lr){3-4}\cmidrule(lr){5-6}
Benchmark & & Correct & Completed & Correct & Completed \\
\midrule
\rowcolor{tablefocus}
TIR & 120 & 87 & 119 & 89 & 117 \\
\rowcolor{tablefocus}
VSTAR & 191 & 168 & 191 & 166 & 188 \\
\rowcolor{tablefocus}
ZoomBench & 845 & 507 & 845 & 521 & 844 \\
HR-Bench 4K & 800 & 662 & 784 & \tbd & \tbd \\
\rowcolor{tablefocus}
HR-Bench 8K & 800 & 658 & 781 & 662 & 761 \\
\bottomrule
\end{tabular}
\end{table}

The clean SFT/SFT HR-Bench 8K evaluation gives $658/800$ ($82.25\%$), with 781 completed rows and judge results for all 781 answers.  The 19 unfinished rows comprise 12 context-length failures, four timeouts from one transient endpoint stall, and three iteration-limit failures; all count as wrong.  Main-only RL with the SFT reader gives $662/800$ ($82.75\%$), with 761 completions.  The $+0.50$-point difference is descriptive: no paired uncertainty analysis is available for this pair.

The older SFT-main observation, $567/800$ with 640 completions, remains excluded from the SFT baseline because 160 rows failed around an unresponsive service and the tile-reader deployment state was unresolved.  The clean replacement above is the SFT reference.  Exact decoded-RGB exclusion and the additional perceptual-overlap sensitivity each leave it unchanged at $658/800\rightarrow658/800$ ($82.25\%\rightarrow82.25\%$, $0.00$ percentage points).  This HR-Bench 8K sensitivity does not establish that other benchmarks are free of overlap.

The HR-Bench 4K comparison uses one SFT run with $662/800$ correct and three pretrained-policy search runs with $627/800$, $638/800$, and $621/800$ correct.  All four use the unseeded protocol and count failures as wrong.  For each gold row, the contrast subtracts the mean correctness of the three pretrained runs from SFT correctness.  Averaging over 800 rows gives $+4.1667$ percentage points.  The 95\% paired percentile bootstrap interval is $[+1.54,+6.75]$ points when resampling the 200 questions with their four option rotations together; resampling the 177 images gives $[+1.54,+6.82]$.  Both analyses use 10,000 resamples.  These intervals describe sampled-example uncertainty conditional on the evaluated runs, not variability across SFT runs.  Restricting each SFT/pretrained pair to commonly completed rows produces different subsets of 762, 764, and 758 rows and gains of $3.28$, $1.70$, and $4.62$ points, respectively.

\subsection{Shared-Actor Four-Response Evaluation}
\label{app:shared_actor}
A separate 100-prompt evaluation samples four responses per prompt and compares SFT with main-only RL, using each policy for both deployed roles.  Pass@1 is mean single-response correctness, and pass@4 is the fraction of prompts with at least one correct response among the four samples.  Historical leakage checks for this set are incomplete.  Table~\ref{tab:shared_actor} preserves the prompt-paired summary statistics; it does not isolate the main policy from changes in the shared tile reader.

\begin{table}[t]
\centering
\caption{Four-response evaluation on 100 prompts. Accuracy differences are percentage points; tool-call differences are calls per trajectory. Intervals are 95\% prompt-paired bootstrap intervals. Shading marks the tool-use contrast. Blue/orange indicate numerical increases/decreases, not benefit or significance.}
\label{tab:shared_actor}
\small
\begin{tabular}{lrrrr}
\toprule
Metric & SFT & RL & $\Delta$ & 95\% CI \\
\midrule
Pass@1 (\%) & 71.25 & 72.50 & \posdelta{+1.25} & $[-3.25,+5.75]$ \\
Pass@4 (\%) & 89.00 & 83.00 & \negdelta{-6.00} & $[-12.0,0.0]$ \\
\rowcolor{tablefocus}
Tool calls & 2.645 & 2.105 & \negdelta{-0.540} & $[-0.87,-0.21]$ \\
\bottomrule
\end{tabular}
\end{table}

\subsection{Joint-Training Diagnostics}
\label{app:joint_dynamics}
The preceding joint stage stays near its warm-start value on internal main-agent validation: two saved evaluations score $0.789$ and $0.793$, and none of 14 paired checkpoint tests has $p\leq0.11$.  This is a lack of resolved change, not evidence of equivalence.  The selected joint RL policy reaches internal main mean@4 of $0.8325$, while sub-agent validation stays at or above $0.985$ on examples drawn from training data.  Over the continuation, training pass@1 rises from $0.4538$ to $0.5205$ while sampled pass@8 stays at $0.7727$ over a prolonged interval.  The prompt population changes between updates, so these aggregates do not measure coverage change on fixed prompts.  Sub-agent validation cannot distinguish task ease from memorization.

\subsection{Role-Matrix Contrasts}
\label{app:role_stats}
Table~\ref{tab:role_matrix_stats} records the paired comparisons behind the two matrices.  The 95\% percentile bootstrap intervals resample 200 underlying questions, keeping all four option rotations together, with 10,000 resamples.  They quantify sampled-question uncertainty conditional on each evaluated pair, not run-to-run variance.  The selected joint RL matrix has not been repeated.  The early mixed RL matrix's second pass scores $82.88/83.38/83.75/85.25\%$ in SFT/SFT, SFT/mixed-sub, RL/SFT, and RL/mixed-sub order.  With the SFT sub-agent fixed, its main-agent difference is $+0.88$ points with interval $[-2.00,+3.75]$.  With the mixed sub-agent fixed, the first pass gives a positive main-agent interval, but the repetition's $+1.88$ points has interval $[-0.25,+4.13]$.  Neither sub-agent contrast resolves a gain in the repetition.  The third pass gives main-agent differences of $+3.62$ points with the SFT reader and $+0.88$ with the early mixed reader.  The main-agent direction is positive in all three passes under both readers.  The third pass has no reported paired uncertainty analysis, so these repeated point estimates do not establish a stable accuracy gain.  They provide no replication of the selected joint RL matrix.

\begin{table*}[t]
\centering
\caption{Paired HR-Bench 4K contrasts in percentage points.  Intervals are 95\% question-paired bootstrap intervals, preserving option rotations. Blue/orange indicate positive/negative estimates. Shading marks the joint-main contrast whose question-paired interval excludes zero.}
\label{tab:role_matrix_stats}
\small
\begin{tabular}{llrl}
\toprule
Matrix & Contrast & $\Delta$ & 95\% CI \\
\midrule
Main-only/early mixed & Main RL $-$ SFT, SFT sub & \posdelta{+0.88} & $[-1.38,+3.25]$ \\
 & Early mixed $-$ SFT sub, SFT main & \negdelta{-1.50} & $[-3.88,+1.00]$ \\
 & Early mixed $-$ SFT sub, RL main & \ensuremath{+0.00} & $[-2.12,+2.25]$ \\
 & Main RL $-$ SFT, early mixed sub & \posdelta{+2.38} & $[+0.25,+4.50]$ \\
Joint RL & Joint RL $-$ SFT sub, SFT main & \posdelta{+1.38} & $[-1.12,+3.88]$ \\
 & Joint RL $-$ SFT main, SFT sub & \negdelta{-2.38} & $[-5.75,+0.75]$ \\
 & Joint RL in both roles $-$ SFT/SFT & \negdelta{-2.50} & $[-5.50,+0.38]$ \\
\rowcolor{tablefocus}
 & Joint RL $-$ SFT main, joint RL sub & \negdelta{-3.88} & $[-6.88,-1.12]$ \\
 & Role interaction & \negdelta{-1.50} & $[-4.62,+1.50]$ \\
\bottomrule
\end{tabular}
\end{table*}

For comparison with the earlier row-level analysis, the joint main-agent drop with the joint sub-agent fixed has exact McNemar $p=0.0022$.  This test compares the counts of examples where only one of the two policies is correct.  Its row-level calculation treats rotations as separate observations, so the question-paired interval in Table~\ref{tab:role_matrix_stats} is the primary uncertainty estimate.  Resampling images instead also gives a negative interval, $[-7.04,-0.89]$ points.  The role interaction remains unresolved under either grouping.

An older exploratory swap kept the SFT main agent and replaced only the sub-agent with the early mixed RL policy.  HR-Bench 4K changed from $662/800$ ($82.8\%$) to $677/800$ ($84.6\%$), and the archived 189-row VSTAR subset changed from $167/189$ ($88.4\%$) to $172/189$ ($91.0\%$).  Neither contrast was significant.  The policy was already in an entropy-uncontrolled regime, its usefulness decisions did not improve, and confidence AUC changed from $0.438$ to $0.573$.  These cells belong to a different evaluation from Table~\ref{tab:role_matrix} and are not results from the later joint RL policy.

\subsection{Joint-Policy Completion and Tool Use}
\label{app:joint_completion}
Table~\ref{tab:joint_completion} separates end-to-end accuracy from accuracy conditional on each policy's completion.  The latter uses different denominators and favors policies that fail on more difficult examples.  With the SFT sub-agent fixed, the main-agent comparison on 716 jointly completed rows is $636/716$ for SFT and $654/716$ for joint RL.  The joint policy gains 18 correct answers there and three on rows only it completes, but loses 40 on the 73 rows completed only by SFT.  The net change is therefore $18+3-40=-19$ correct answers.  Deploying joint RL in both roles similarly gains 10 answers on 719 jointly completed rows and one on its exclusively completed rows, but loses 31 when it fails, for a net change of $-20$.

\begin{table}[t]
\centering
\caption{Completion sensitivity and tool use for the joint RL role matrix.  Conditional accuracy uses only that cell's completed rows; all end-to-end scores in Table~\ref{tab:joint_v4_role_matrix} use 800 rows. Shading identifies the joint-main arms with longer tool-call tails.}
\label{tab:joint_completion}
\small
\begin{tabular}{llrrr}
\toprule
Main & Sub & Correct/completed & Mean calls & Rows with $\geq10$ calls \\
\midrule
SFT & SFT & 676/789 & 2.22 & 18 \\
SFT & Joint RL & 687/790 & 2.11 & 14 \\
\rowcolor{tablefocus}
Joint RL & SFT & 657/720 & 4.35 & 107 \\
\rowcolor{tablefocus}
Joint RL & Joint RL & 656/722 & 4.37 & 105 \\
\bottomrule
\end{tabular}
\end{table}

With SFT as main, swapping the tile reader gives $672/781$ versus $685/781$ correct on commonly completed rows, preserving the small positive direction.  The full-row gain remains unresolved.  With joint RL as main, the SFT-sub and joint-sub cells incur 36/43 iteration-limit failures and 40/35 context-length failures, respectively.  Their failed trajectories average $12.21/13.37$ tool calls, compared with $3.48/3.40$ among completed trajectories.  The 80/78 failed rows span 32/33 underlying questions, including 11/10 questions failing all four rotations.  The context errors hit the 32{,}768-token service cap; some record only a lower bound on input length, so a full token-length distribution is unavailable.  These diagnostics connect the accuracy loss to long trajectories and termination; the question-paired intervals in Table~\ref{tab:role_matrix_stats} account for dependence among rotations.

\subsection{Oracle-Evidence Diagnostic}
A separate zoom-only RL diagnostic compares two 152-update arms with 76 crop-injected training questions, 76 non-injected training questions, and 97 held-out questions.  The intervention replaces the model-selected crop with the gold ROI crop on the injected subset.  Across 296 paired update--question observations on that subset, accuracy rises from $75.17\%$ to $84.97\%$ ($+9.80$ points).  The non-injected subset gains $1.67$ points across 284 such observations, falling to $0.88$ points after excluding trajectories that hit the rollout or turn budget; the held-out estimate is also $+1.67$ points across 285 observations.  These counts include repeated questions at different updates and are not independent-question sample sizes.  The larger effect on injected examples supports evidence consumption as a diagnostic, while the transfer estimates do not establish improved search.  Because the intervention supplies gold crops through \texttt{zoom\_in}, it does not measure the causal contribution of returned grid evidence.

\subsection{Forced-Grid and Valid-Grid Diagnostics}
\label{app:grid_diagnostics}
A separate exploratory main-agent RL arm begins with the VPS SFT policy and requires one fixed $3\times3$ \texttt{grid\_search} before answering.  Eight responses to each question share one cached, greedy tile-report prefix, and subsequent zoom is disabled.  The tile report enters the context as an observation; the answer receives the main-agent reward.  Across six successive forced-grid evaluations, accuracy on the same 200 internal inputs rises from $44.0\%$ to a maximum of $67.5\%$, while unfinished trajectories fall from $47.0\%$ to $9.0\%$ at that maximum.  The last saved forced-grid policy scores $65.5\%$ with $8.0\%$ unfinished under the forced protocol.  When placed back in the normal grid-and-zoom environment, it scores $67.5\%$, uses a successful grid call on $9.5\%$ of inputs, uses zoom on $44.5\%$, and has no unfinished outputs in that evaluation.

The continuation begins from that last forced-grid policy and restores free tool choice.  Its training mixture contains 1,997 main-agent questions and 499 sub-agent tile prompts.  For a correct main-agent answer, the reward is $0.95$ without a successful grid call and $1.00$ with at least one successful grid call; incorrect answers receive zero.  Table~\ref{tab:valid_grid_diagnostic} reports successive greedy evaluations on the same inputs.  The final continuation increases successful grid use from $9.5\%$ to $21.0\%$ while accuracy falls to $64.5\%$ and unfinished outputs rise to $10.0\%$.  There is no synchronized continuation without the grid reward from the same initialization, and the change from forced-grid training also changes the agent loop and role mixture.  This diagnostic therefore describes policy behavior rather than an isolated causal reward effect.

\begin{table}[t]
\centering
\caption{Exploratory default-environment evaluations on 200 RL-held-out internal inputs (greedy, one response each).  The inputs are disjoint from the current RL training pool by sample identity and image content, but 196 exact main-agent rows appeared in the earlier SFT training pool.  ``Grid'' is the rate of at least one successful grid call. Shading marks the final continuation, where increased grid use accompanies lower accuracy.}
\label{tab:valid_grid_diagnostic}
\small
\begin{tabular}{lrrrr}
\toprule
Policy & Accuracy & Grid & Zoom & Unfinished \\
\midrule
Forced-grid warm start & 67.5\% & 9.5\% & 44.5\% & 0.0\% \\
Continuation 1 & 71.5\% & 10.0\% & 47.5\% & 7.0\% \\
Continuation 2 & 69.5\% & 14.0\% & 47.5\% & 5.0\% \\
Continuation 3 & 67.5\% & 15.0\% & 49.0\% & 8.0\% \\
\rowcolor{tablefocus}
Continuation 4 & 64.5\% & 21.0\% & 47.5\% & 10.0\% \\
\bottomrule
\end{tabular}
\end{table}

The current main-only RL comparison in Table~\ref{tab:role_matrix} used 1,997 main-agent training prompts after removing 136 inputs that overlapped its 200-input validation set.  Each optimizer update used eight prompts and eight rollouts per prompt, one policy update epoch, and a batch of four trajectories per optimizer mini-batch.  The resolved actor used LoRA rank and scaling 32/32, AdamW at a constant $10^{-5}$ learning rate, weight decay $0.01$, gradient clipping $1.0$, a KL-loss coefficient of $0.001$, and entropy coefficient $-0.01$; zero-variance group filtering was disabled.  Training sampled at temperature $1.0$ and top-$p$ $0.95$, while greedy validation used one response per prompt.  The tool-enabled training loop allowed up to 16 assistant and 16 user turns, a 32,768-token model window, and both \texttt{grid\_search} and \texttt{zoom\_in}.  These settings describe that main-only RL arm; the joint-training lineage and the training-free harness have separate configurations.

\subsection{Joint-System Gradient and the Implemented Truncation}

Let $\theta_O$ and $\theta_S$ parameterize the main and sub roles, respectively.  The derivation assumes stochastic behavior policies with retained log probabilities, fixed tool transitions and aggregation, and no separately learned selector; a learned router would contribute its own score term, while greedy or beam-search traces are not unmodified on-policy samples.  For a system trajectory $\tau$, these assumptions give
\begin{align}
\nabla_{\theta_O}J_G
&=\mathbb E\!\left[(R_G-b_O)\sum_{u\in\mathcal I_O(\tau)}
\nabla_{\theta_O}\log\pi_{\theta_O}^{\mathrm{main}}(a_u\mid h_u)\right],\\
\nabla_{\theta_S}J_G
&=\mathbb E\!\left[(R_G-b_S)\sum_{v\in\mathcal I_S(\tau)}
\nabla_{\theta_S}\log\pi_{\theta_S}^{\mathrm{sub}}(a_v\mid h_v)\right].
\end{align}
The histories are the pre-action histories of the corresponding calls; copied grid-report tokens in the main context are observations and cannot substitute for the original sub-agent samples or their behavior-policy log probabilities.  Conditional advantages can replace the terminal returns, but a local tile history integrates over the other parallel reports and the downstream main trajectory, making the sub-agent estimator high variance.

For the direct local objective, let $s\sim\nu$ be a tile prompt drawn from a distribution held fixed during the actor update and $z\sim\pi_{\theta_S}^{\mathrm{sub}}(\cdot\mid s)$.  Then
\begin{equation}
J_{L,\mathrm{dir}}^{\nu}
=\mathbb E_{s,z}[R_L(s,z)],\qquad
\nabla_{\theta_O}J_{L,\mathrm{dir}}^{\nu}=0.
\end{equation}
Its sub-agent gradient contains only the score of $z$.  If $\nu$ is refreshed from main-agent grid calls, this statement is exact for the frozen per-update replay objective but omits the grid-occupancy derivative of the corresponding on-policy objective.  A fully on-policy grid-local objective would additionally credit main actions that determine whether and which tiles exist, and sub-agent actions that affect later local returns.

Equation~\ref{eq:joint_objective} retains the main/global and direct-sub/local terms while dropping system-sub/global credit.  An external tile pool has no main-policy occupancy term; the grid-derived replay variant also omits the occupancy derivative of its corresponding on-policy objective.  Token masks implement this truncation: main-generated tokens receive $A_G$, independently resampled local-response tokens receive $A_L$, and prompts, copied tool results, and nested system sub-agent responses receive neither loss.  When parameters are shared, the resulting update remains coupled through the weights even though the score terms are masked.  Finite-group GRPO normalization---especially a mean and standard deviation computed using the current sample---is not generally an unbiased conditional advantage; clipping and repeated epochs add further surrogate approximations.  Role groups must therefore be formed separately from identical prompts, and the behavior snapshot must be refreshed after shared-weight updates.

\begin{samepage}
\subsection{Paired RL Pseudocode}
\begin{enumerate}
    \item Sample $K$ main trajectories for each source prompt.
    \item Draw sub-agent prompts from the fixed external tile dataset.  For the grid-derived replay variant only, select a tile from a valid original-image grid call using a fixed geometry-only sampler and freeze the resulting prompt batch.
    \item Sample $K$ sub-agent responses per identical tile prompt, independently of the main-prompt groups.
    \item Compute main and sub rewards and normalize advantages within separate group identifiers.
    \item Mask copied tool results and nested system sub-agent outputs from the global loss; apply the local loss only to the separately sampled tile responses.
    \item When configured, filter zero-variance groups; then apply task weights and perform one paired multi-task actor update before refreshing the behavior snapshot.
\end{enumerate}
\end{samepage}

\subsection{Completion-Aware Reward Pilot}
\label{app:hardfail}
We compare two 40-update main-only runs from the same SFT initialization, with 1,997 training prompts, eight prompts per update, eight rollouts per prompt, temperature $1.0$, and a 24,576-token response limit.  The seed, optimizer, hardware, and judge services are shared.  The intervention changes only the reward entry point: an unfinished trajectory receives zero terminal correctness reward, while the control retains the existing answer reward.  Unfinished samples remain in the training batch, and response-token masking is unchanged.  Evaluation counts unfinished trajectories as wrong in both arms.

Table~\ref{tab:hardfail} reports all five evaluations on the same 200 internal prompts, one response per prompt.  These prompts are disjoint from the current RL training pool but exposed during SFT (196 training and four validation rows).  Even the initial evaluations of the same SFT weights differ, illustrating rollout variation.  At the final evaluation, the intervention scores $151/200$ versus $139/200$, with 24 intervention-only and 12 control-only successes (two-sided exact McNemar $p=0.065$).  Unfinished counts are 14 versus 22, with seven intervention-only and 15 control-only failures ($p=0.134$).  The positive endpoint is a pilot signal: intermediate accuracy differences reverse, the paired tests do not resolve a difference at the 5\% level, and external transfer has not been evaluated.  Conditional accuracy at the endpoint is $151/186=81.2\%$ versus $139/178=78.1\%$; these denominators select different completed subsets.

\begin{table}[t]
\centering
\caption{Completion-aware reward pilot on 200 SFT-exposed internal prompts. All values are percentages; the initial evaluation precedes training, followed by four equally spaced evaluations. Shading marks the final evaluation; it does not indicate a resolved or externally validated gain.}
\label{tab:hardfail}
\small
\begin{tabular}{lrrrr}
\toprule
& \multicolumn{2}{c}{Accuracy} & \multicolumn{2}{c}{Unfinished} \\
\cmidrule(lr){2-3}\cmidrule(lr){4-5}
Evaluation & Intervention & Control & Intervention & Control \\
\midrule
Initial & 71.0 & 69.0 & 15.5 & 15.0 \\
Quarter & 69.5 & 69.5 & 9.5 & 13.0 \\
Half & 69.5 & 71.5 & 9.5 & 10.0 \\
Three-quarter & 72.0 & 73.5 & 8.0 & 8.5 \\
\rowcolor{tablefocus}
Final & 75.5 & 69.5 & 7.0 & 11.0 \\
\bottomrule
\end{tabular}
\end{table}

Each arm generates 2,560 training trajectories.  The intervention marks 105 as unfinished and applies zero correctness reward to all 105; only a few scores change, reducing average correctness from $72.34\%$ to $72.23\%$.  The control records 109 unfinished trajectories and $71.80\%$ correctness.  Broad truncation indicators occur on $49.22\%$ and $51.25\%$ of trajectories, respectively, far more often than the $4.10\%$ and $4.26\%$ unfinished rates.  Truncation and semantic non-completion must therefore remain separate diagnostics.

\subsection{Utilization of Relative-Reward Groups}
\label{app:group_utilization}
The selected main-only policy uses neither offline difficulty filtering nor online zero-variance group filtering.  Across its 800 training prompt groups, 284 ($35.50\%$) are all-correct, 38 ($4.75\%$) all-wrong, and 478 ($59.75\%$) mixed.  Thus $40.25\%$ have zero correctness advantage, although entropy or KL regularization can still act.  The substantial mixed fraction does not establish saturation of the SFT policy.

An earlier main-only run with online filtering retains 688 groups and evicts 1,048 ($60.4\%$ of groups considered).  A separate sub-only recipe filters 2,133 prompts offline to 1,358, removing 759 all-correct and 16 all-wrong prompts.  These runs change other settings as well and do not isolate either a wall-clock speedup or a held-out accuracy benefit from filtering.  Offline removal changes the training distribution; online rejection adds generation work and changes the distribution of reported surviving groups.  Neither should be conflated with the unfiltered main-only policy or the joint warm-start lineage.

\subsection{Data Snapshot and Annotation Audit}
\label{app:data_audit}
The evaluated SFT snapshot contains 6,464 input examples (2,197 main and 4,267 sub), split at runtime into 6,336 training and 128 validation records.  A separate SFT test set and the role/source composition of this split are not documented.  Conversion yields 2,133 main-agent records; removing 136 RL-validation overlaps leaves 1,997 training prompts, comprising 1,057 ZwZ and 940 VisualProbe examples.  The reason-level accounting of the 64 conversion losses remains unavailable.  Current RL train/validation checks use sample identity, group identity, original-image path, and normalized RGB content.  The external-benchmark audit covers nine direct SFT/RL source pools and all five benchmark splits.  It separates exact byte identity, exact decoded-RGB identity, and perceptual near-duplicate candidates.  Table~\ref{tab:external_overlap} reports benchmark-stratified unique-image counts; source pools can share images, so their counts must not be summed as a union.

For the ROI audit, two annotators each received 500 images, with no double annotation.  The sample contains 500 VisualProbe and 500 ZwZ examples; 491 and 445 receive usable labels, while nine and 55 are skipped.  Instructions request tight boxes around answer evidence, allow multiple boxes when multiple objects matter, and ask annotators to flag ambiguous, mismatched, or overly broad questions.  Examples are skipped when no ROI can be marked.  Consequently, annotation agreement is not measured.  Further administration and adjudication details remain undocumented.

The tile-labeling log records 6,000 candidates and 5,997 successful labels.  The hint-free verifier checks 1,800 examples and rejects 72; a separate rewrite audit records 235 model rewrites and 16 fallback rewrites.  These stage totals do not yet reconcile to a single final export manifest, so they are not presented as a complete filtering flow or as the final SFT sample count.

\paragraph{External-Benchmark Image Overlap}
Exact decoded-RGB identity compares pixel content rather than encoded file bytes; it is the audit's exact image-content check, separate from perceptual similarity.  Exact matches in the audited pools are confined to ZoomBench.  The audit's VSTAR image inventory includes 238 entries, while answer accuracy uses the 191 gold-labeled rows.  Image-inventory counts and scored-row denominators therefore describe different populations.

\begin{table}[t]
\centering
\caption{Unique benchmark images with direct training-source overlap. Each cell reports exact byte / exact decoded-RGB matches separately. Counts are per pool, not cumulative checkpoint exposure; all nonzero exact matches are on ZoomBench.}
\label{tab:external_overlap}
\small
\setlength{\tabcolsep}{4pt}
\begin{tabular}{lrrrrr}
\toprule
Source pool & TIR & VSTAR & ZoomBench & HR-4K & HR-8K \\
\midrule
SFT main & 0/0 & 0/0 & 0/0 & 0/0 & 0/0 \\
SFT sub & 0/0 & 0/0 & 3/11 & 0/0 & 0/0 \\
Main-only RL & 0/0 & 0/0 & 4/8 & 0/0 & 0/0 \\
Early mixed RL & 0/0 & 0/0 & 3/10 & 0/0 & 0/0 \\
Joint stage 1 & 0/0 & 0/0 & 3/8 & 0/0 & 0/0 \\
Joint stage 2 & 0/0 & 0/0 & 4/10 & 0/0 & 0/0 \\
Joint stage 3 & 0/0 & 0/0 & 8/23 & 0/0 & 0/0 \\
Selected joint stage & 0/0 & 0/0 & 4/11 & 0/0 & 0/0 \\
Valid-grid mixture & 0/0 & 0/0 & 6/14 & 0/0 & 0/0 \\
\bottomrule
\end{tabular}
\end{table}

The perceptual sensitivity uses 64-bit pHash with Hamming distance at most four, excluding exact decoded-RGB matches from the candidate list.  All returned candidates have distance zero.  Deduplicating across source-pool records yields ten unique image pairs: nine on ZoomBench and one on TIR.  Their $64\times64$ EXIF-normalized RGB correlations are at least $0.9999$, consistent with resized or re-encoded versions.  These are near-duplicate candidates for sensitivity analysis, not proven contamination.  No exact or perceptual match is detected on VSTAR, HR-Bench 4K, or HR-Bench 8K under these checks.

\paragraph{Cumulative Exposure and Exclusion Sensitivity}
Exclusion accumulates exposure across both deployed roles and their ancestors: both SFT sources, the main-only RL pool where applicable, and the selected joint policy's warm-start stages.  Exact decontamination removes rows matching this union by decoded RGB; perceptual sensitivity additionally removes near-duplicate candidates.  Remaining failed or unfinished rows count as wrong.

Table~\ref{tab:overlap_sensitivity} gives the cross-benchmark sensitivities.  On ZoomBench, exact exclusion removes 11 SFT and 19 main-only RL rows; adding perceptual candidates removes 20 rows per lineage.  Different lineage exposures yield different excluded sets, so these are not matched policy comparisons.  All HR-Bench 4K cells in Tables~\ref{tab:role_matrix} and~\ref{tab:joint_v4_role_matrix} retain their original numerators and 800-row denominators under both exclusions ($0.00$-point changes).  Across the 25 audited evaluations, the largest absolute accuracy shift is $0.28$ points.  This sensitivity covers the audited image matches, not all forms of training exposure.

\begin{table}[t]
\centering
\caption{Lineage-specific exclusion sensitivity. Cells show correct/total (accuracy, \%); exclusion cells also give $\Delta$ from the original accuracy in percentage points. The pHash column includes exact-RGB exclusions. SFT and main-only RL both use the SFT reader; HR-4K here is the historical SFT run. All counts retain failures as wrong.}
\label{tab:overlap_sensitivity}
\small
\setlength{\tabcolsep}{4pt}
\begin{tabular}{llrrr}
\toprule
Benchmark & Main & Original & Exact RGB & + pHash \\
\midrule
TIR & SFT & 87/120 (72.50\%) & \shortstack[r]{87/120 (72.50\%)\\$\Delta=+0.00$} & \shortstack[r]{86/119 (72.27\%)\\$\Delta=-0.23$} \\[3pt]
TIR & Main-only RL & 89/120 (74.17\%) & \shortstack[r]{89/120 (74.17\%)\\$\Delta=+0.00$} & \shortstack[r]{88/119 (73.95\%)\\$\Delta=-0.22$} \\[3pt]
VSTAR & SFT & 168/191 (87.96\%) & \shortstack[r]{168/191 (87.96\%)\\$\Delta=+0.00$} & \shortstack[r]{168/191 (87.96\%)\\$\Delta=+0.00$} \\[3pt]
VSTAR & Main-only RL & 166/191 (86.91\%) & \shortstack[r]{166/191 (86.91\%)\\$\Delta=+0.00$} & \shortstack[r]{166/191 (86.91\%)\\$\Delta=+0.00$} \\[3pt]
ZoomBench & SFT & 507/845 (60.00\%) & \shortstack[r]{502/834 (60.19\%)\\$\Delta=+0.19$} & \shortstack[r]{495/825 (60.00\%)\\$\Delta=+0.00$} \\[3pt]
ZoomBench & Main-only RL & 521/845 (61.66\%) & \shortstack[r]{507/826 (61.38\%)\\$\Delta=-0.28$} & \shortstack[r]{507/825 (61.45\%)\\$\Delta=-0.20$} \\[3pt]
HR-4K & SFT & 662/800 (82.75\%) & \shortstack[r]{662/800 (82.75\%)\\$\Delta=+0.00$} & \shortstack[r]{662/800 (82.75\%)\\$\Delta=+0.00$} \\[3pt]
HR-8K & SFT & 658/800 (82.25\%) & \shortstack[r]{658/800 (82.25\%)\\$\Delta=+0.00$} & \shortstack[r]{658/800 (82.25\%)\\$\Delta=+0.00$} \\[3pt]
HR-8K & Main-only RL & 662/800 (82.75\%) & \shortstack[r]{662/800 (82.75\%)\\$\Delta=+0.00$} & \shortstack[r]{662/800 (82.75\%)\\$\Delta=+0.00$} \\[3pt]
\bottomrule
\end{tabular}
\end{table}

\subsection{Recovered Training Configuration}
\label{app:training_config}
SFT starts from Qwen3.5-4B with LoRA rank/scaling $32/64$ on all linear modules, a trainable vision tower, and a frozen aligner.  It uses two epochs, a $10^{-4}$ learning rate with cosine scheduling, bfloat16 arithmetic, a 32,768-token maximum length, and seed 42.  The main-only RL configuration described above uses four A800 GPUs with FSDP2, a 3,584-token prompt limit, and 24,576-token response and tool-response limits.  Its independently evaluated tile-reader service is fixed to SFT where indicated; sharing actor parameters during training does not itself freeze that deployed role.

The selected joint stage uses 893 training and 264 validation examples, batches of 16 prompts, eight responses per prompt, optimizer mini-batches of two trajectories, and LoRA on seven projection modules.  Validation uses four responses per prompt at temperature $1.0$.  It follows multiple warm-started stages; it does not share the main-only run's initialization or data mixture.  The local prompts come from a fixed external tile dataset.  Group identifiers distinguish role, sample, and tile, so only identical prompts participate in each relative-reward normalization.

Historical joint reward code, the exact scalar range and error fallback, numerical local-loss weighting, and loss-aggregation details have not been recovered.  The preserved description establishes the structured-output gate and scalar local judge, but current reward code cannot establish the historical implementation.  Exact historical tool and judge prompts, old service revisions, and complete visual-token and resizing policies are also unavailable, limiting exact reproduction of those runs.